# Near-Infrared Coloring via a Contrast-Preserving Mapping Model

Chang-Hwan Son and Xiao-Ping Zhang

*Abstract*—Near-infrared gray images captured together with corresponding visible color images have recently proven useful for image restoration and classification. This paper introduces a new coloring method to add colors to near-infrared gray images based on a contrast-preserving mapping model. A naive coloring method directly adds the colors from the visible color image to the near-infrared gray image; however, this method results in an unrealistic image because of the discrepancies in brightness and image structure between the captured near-infrared gray image and the visible color image. To solve the discrepancy problem, first we present a new contrast-preserving mapping model to create a new near-infrared gray image with a similar appearance in the luminance plane to the visible color image, while preserving the contrast and details of the captured near-infrared gray image. Then based on the proposed contrast-preserving mapping model, we develop a method to derive realistic colors that can be added to the newly created near-infrared gray image. Experimental results show that the proposed method can not only preserve the local contrasts and details of the captured near-infrared gray image, but transfers the realistic colors from the visible color image to the newly created near-infrared gray image. Experimental results also show that the proposed approach can be applied to near-infrared denoising.

## I. INTRODUCTION

Near-infrared imaging was developed to consecutively capture near-infrared gray images and visible color images [1]. In general, a hot mirror is used to prevent the near-infrared part of the electromagnetic spectrum, ranging from 750 to 1400 nm, from reaching sensitive CMOS sensors and contaminating the visible color images. However, if the hot mirror is replaced with a piece of clear glass and a pair of lens-mounted filters to block or pass the near-infrared region, a visible color image and its corresponding near-infrared gray image can be captured. This pair of images often has different pixel information for the same scene. The resulting redundant information can be helpful in certain tasks, including image denoising [2], deblurring [3], dehazing [4], and classification [5].

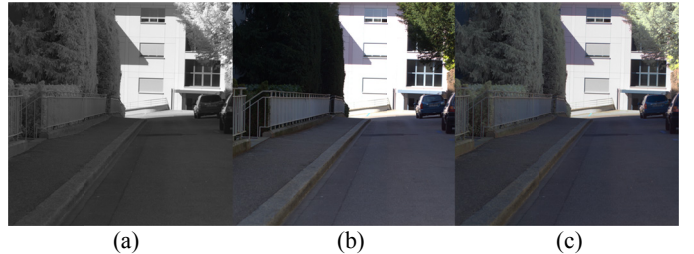

(a)                    (b)                    (c)

Fig. 1. (a) Near-infrared gray image, (b) visible color image, and (c) colored infrared image.

### A. Motivation

The objective of this paper is to provide a new way to color a near-infrared gray image. This is ostensibly easy to accomplish because visible color images are available. It is tempting to use a naive coloring method that merely combines the chrominance planes from the visible color image with the near-infrared gray image in an opponent color space. However, discrepancies can emerge between the near-infrared gray image and the visible color image. In other words, the brightness and image structure of near-infrared gray images can differ from those of the visible color images. An example is provided in Fig. 1. In Fig. 1(a), the near-infrared gray image shows the leaves on the tree and the lines on the wall. However, these are absent in the visible color image in Fig. 1(b). This is caused by the low dynamic range of the camera that leads to pixel saturation in the dark and bright regions. Therefore, a naive coloring method to directly add colors from the visible color image to the near-infrared gray image results in unnatural image colors, as shown in Fig. 1(c). In particular, the colors of the tree in the colored near-infrared image are unnatural, in contrast to those of the tree in the visible color image. Therefore, this paper focuses on a method for solving this discrepancy so that the natural colors from the visible color image can be transferred to a near-infrared gray image without any loss of detail and contrast.

### B. Contribution

● A new method for transferring the colors from a visible color image to a near-infrared gray image is presented. This method first involves modeling a contrast-preserving linear mapping between the two images. This mapping model is used to generate a new near-infrared gray image and then to correct the chrominance distribution of the visible color image, thereby transferring realistic colors to the newly created near-infrared gray image. Moreover, the proposed mapping model can resolve the discrepancy between the

This work was supported by the National Research Foundation of Korea (NRF-2013R1A1A2061165), the Natural Sciences and Engineering Research Council of Canada (NSERC) under Grant RGPIN239031

C.-H. Son is with the College of Information and Communication Engineering, Sungkyunkwan University, Suwon, 440-746, South Korea (E-mail: changhwan76.son@gmail.com.)

X.-P. Zhang is with the Department of Electrical and Computer Engineering, Ryerson University, ON M5B2K3 Canada (E-mail: xzhang@ryerson.ca).

near-infrared gray image and visible color image.
- A detail layer transfer method based on the detail difference constraint is introduced to enhance the detail layer of the newly created near-infrared gray image. The detail difference constraint forces the gradients of the detail layer of the newly created near-infrared gray image to be closer to those of the captured near-infrared gray image. By adopting this detail difference constraint, detail description and the amount of visual information can be improved.
- Conventional methods [1–4] focus on fusing images to increase visual information or edge representation. However, the proposed method is concerned with transferring the colors from the visible color image to the near-infrared gray image without any loss of detail and contrast. The differences between the conventional and proposed methods are discussed in this paper. The experimental results confirm that the proposed method is more effective at expressing local color contrast and detail than conventional methods.
- In dim lighting conditions, captured visible color images can contain significant noise. In this case, the question arises as to whether the proposed method effectively works on the captured image pair of a noisy visible color image and the near-infrared gray image. The experimental results show that the proposed approach used for the near-infrared coloring can also be applied to image pairs captured under dim lighting conditions. Moreover, it is discussed how the proposed near-infrared denoising differs from conventional image-pair-based fusion [2,3,6,7,8,9,10].

## II. RELATED WORK

### A. Regularization Approach

If the discrepancy problem can be ignored, a naive coloring method that involves combining the chrominance planes of the visible color image with the near-infrared gray image can be an adequate solution. However, in most cases, the discrepancy problem cannot be ignored, and thus, more sophisticated techniques are required. One solution is to adopt a regularization approach [3,6,7] that has been widely used for image-pair-based restoration. The regularization term can be modeled as follows:

$$\min_{\mathbf{x}} \left\{ \mu_g \cdot \left\| \mathbf{x} - \mathbf{x}^{v_l} \right\|^2 + \sum_{j=1}^{2} \left| (\mathbf{x} \otimes f^j) - (\mathbf{x}^{nir} \otimes f^j) \right|^\gamma \right\} \quad (1)$$

where $\mathbf{x}^{v_l}$ and $\mathbf{x}^{nir}$ indicate the luminance plane for the visible color and near-infrared gray images, respectively, and $f$ denotes the horizontal or vertical derivative filter. In (1), the first term, i.e., the data-fidelity term, indicates that the unknown luminance image to be estimated is similar to the luminance plane for the visible color image. However, to improve the edge representation, a regularization term is needed that is provided by the second term in (1). This constraint forces the edges from the unknown luminance plane to be close to those in the near-infrared gray image. In (1), $\gamma$ is a constant value that controls the sparsity [8]. In general, the value of $\gamma$ is less than one. Given an estimated luminance plane, the colors can be taken directly from the visible color image. In other words, the estimated luminance plane can be combined with the chrominance planes from the visible color image.

### B. Multiresolution Approach

Another approach is to apply multiresolution techniques [1,9,10] widely used for image fusion. The wavelet transform is a well-known multiresolution representation, using which, the luminance plane for the visible color image can be combined with the luminance plane for the near-infrared gray image as follows:

$$\Theta(\mathbf{x}) = \begin{cases} \omega_h \Theta(\mathbf{x}^{v_l}) + (1 - \omega_h) \Theta(\mathbf{x}^{nir}) & \text{for LL subband} \\ \text{MAX}(\Theta(\mathbf{x}^{v_l}), \Theta(\mathbf{x}^{nir})) & \text{for other subbands} \end{cases} \quad (2)$$

where $\Theta(\mathbf{x}^{v_l})$ and $\Theta(\mathbf{x}^{nir})$ indicate the wavelet coefficients of the luminance planes for the visible color and near-infrared gray images, respectively, and MAX denotes the function that returns the largest value of a set of values. The above equation shows that the wavelet coefficients of the luminance plane for the visible color image are linearly mixed with those of the near-infrared gray image for the lowest frequency subband. However, for other subbands, a larger wavelet coefficient between $\Theta(\mathbf{x}^{v_l})$ and $\Theta(\mathbf{x}^{nir})$ is selected to increase the details of the colorized gray image. Given the created luminance plane, according to (2), its colors can be taken directly from the visible color image.

### C. Statistical Approach

In [11], mean and variance, which are the representative statistics of natural images, are used to transfer the color appearance of a reference image to that of a target image. This approach can also be adopted to solve the near-infrared coloring problem. In other words, the local mean and variance of the visible color image luminance plane can be changed according to the local mean and variance of the near-infrared gray image as follows:

$$\mathbf{x}_i^{\theta_l} = \mu_i^{nir} + (\mathbf{x}_i^{v_l} - \mu_i^{v_l}) \cdot \frac{\sigma_i^{nir}}{\sigma_i^{v_l}} \quad \text{and} \quad \mathbf{x}_i^{\theta_c} = \mathbf{x}_i^{v_c} \cdot \frac{\sigma_i^{nir}}{\sigma_i^{v_l}} \quad (3)$$

where subscript $i$ indicates pixel location, $\mu_i$ and $\sigma_i$ denote the mean and variance of the local patch centered at the $i$th pixel location, respectively, and $\mathbf{x}_i^{\theta_l}$ and $\mathbf{x}_i^{\theta_c}$ indicate the $i$th pixel values of the luminance and chrominance planes of the colorized image, respectively. The above equation tells us that the luminance plane of the visible color image $\mathbf{x}_i^{v_l}$ can be scaled according to the variance ratio $\sigma_i^{nir}/\sigma_i^{v_l}$. In addition, the new chrominance plane $\mathbf{x}_i^{\theta_c}$ can also be created by scaling the chrominance planes $\mathbf{x}_i^{v_c}$ of the visible color image by the variance ratio.

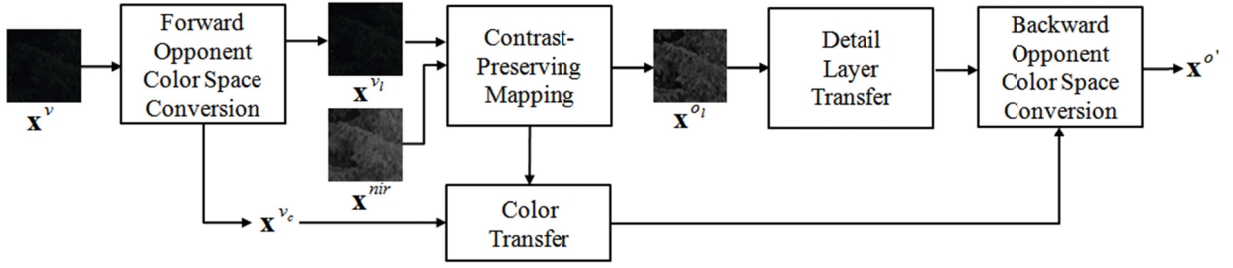

Fig. 2. Block-diagram of proposed method.

## III. PROPOSED NEAR-INFRARED COLORING

Figure 2 shows a block-diagram of the proposed method. First, the visible color image $\mathbf{x}^v$ is separated into the luminance and chrominance planes $\mathbf{x}^{v_l}$ and $\mathbf{x}^{v_c}$ through a forward opponent color space conversion, and then the proposed contrast-preserving mapping is applied to the image pair consisting of the near-infrared gray image $\mathbf{x}^{nir}$ and the luminance plane of the visible color image $\mathbf{x}^{v_l}$. The details and visual information of the newly created luminance plane $\mathbf{x}^{o_l}$ via the contrast-preserving mapping are further enhanced by the detail layer transfer. The relation between $\mathbf{x}^{nir}$ and $\mathbf{x}^{v_l}$ established via the contrast-preserving mapping is used for the color transfer, which changes the color distribution of the chrominance plane $\mathbf{x}^{v_c}$. Next, the detail-enhanced version of the newly created luminance plane is combined with the modified chrominance plane. Finally, the colored output image $\mathbf{x}^{o'}$ is obtained via the backward opponent color space conversion.

### A. Proposed Approach

The central idea of the proposed near-infrared coloring method is the contrast-preserving mapping model. As shown in Fig. 1, the discrepancy problem results in unnatural colors. To handle this issue, a contrast-preserving mapping model is proposed. The first role of this model is to create a new near-infrared gray image. As shown in Fig. 2, the newly created near-infrared gray image $\mathbf{x}^{o_l}$ can preserve the detail and local contrast of the near-infrared gray image $\mathbf{x}^{nir}$. The input visible and near-infrared patches, as shown in Fig. 2, are extracted from the tree region of Fig. 1. However, the newly created near-infrared gray image $\mathbf{x}^{o_l}$ is different from the captured near-infrared gray image $\mathbf{x}^{nir}$. The second role is to determine the mapping relation between the near-infrared gray image $\mathbf{x}^{nir}$ and the visible color image $\mathbf{x}^{v_l}$. A critical point of the proposed coloring method is that it adds unknown colors to the newly created near-infrared gray image $\mathbf{x}^{o_l}$, not the captured near-infrared gray image $\mathbf{x}^{nir}$. To estimate the unknown colors of the newly created near-infrared gray image, the mapping relation obtained using the contrast-preserving mapping is utilized. In other words, the mapping relation provides a new color transfer model to predict the unknown colors from the chrominance images $\mathbf{x}^{v_c}$ of the visible color image.

### B. Contrast-Preserving Linear Mapping

To transfer the colors from the visible color image to the near-infrared gray image without any loss of detail, a contrast-preserving mapping is first needed. The objective of the proposed contrast-preserving mapping is to find the relation between the luminance plane for the visible color image and near-infrared gray image. Based on this mapping relation, another near-infrared gray image is generated that preserves the local contrast. Furthermore, the colors corresponding to the newly created near-infrared gray image can be estimated. The proposed contrast-preserving mapping model can be formulated as follows:

$$\min_{\boldsymbol{\alpha}} \left\| \mathbf{W}^{1/2}(\mathbf{p}_i - \mathbf{Q}_i \boldsymbol{\alpha}_i) \right\|^2 + \mu_c \left\| \boldsymbol{\alpha}_i - \boldsymbol{\alpha}_i^0 \right\|^2 \tag{4}$$

where $\mathbf{p}_i$ and $\mathbf{Q}_i$ contain the pixel values of the extracted patches from the luminance planes of the visible color and near-infrared images at the $i$th pixel location, respectively. A decorrelated color space [11] is used to generate these luminance planes. Other opponent color spaces, e.g., CIELAB or YCbCr [1], could be considered. Assuming that the extracted patch has an odd size $m \times m$, $\mathbf{p}_i$ and $\mathbf{Q}_i$ can be defined as follows:

$$\mathbf{p}_i = \mathbf{R}_i \mathbf{x}^{v_l} = \left[ \mathbf{x}^{v_l}_{i-m/2}, \mathbf{x}^{v_l}_{i-m/2+1}, \ldots, \mathbf{x}^{v_l}_i, \ldots, \mathbf{x}^{v_l}_{i+m/2-1}, \mathbf{x}^{v_l}_{i+m/2} \right]^T \tag{5}$$

$$\mathbf{Q}_i = [\mathbf{R}_i \mathbf{x}^{nir} \ \mathbf{1}] = \left[ \left[ \mathbf{x}^{nir}_{i-m/2}, \mathbf{x}^{nir}_{i-m/2+1}, \ldots, \mathbf{x}^{nir}_i, \ldots, \mathbf{x}^{nir}_{i+m/2-1}, \mathbf{x}^{nir}_{i+m/2} \right]^T \ \mathbf{1} \right] \tag{6}$$

where $\mathbf{R}_i$ is a matrix that extracts the patch at the $i$th pixel location from an image [12] and $T$ denotes the transpose operator. In (6), $\mathbf{1}$ indicates the column vector filled with one. If $\mathbf{x}^{v_l}$ and $\mathbf{x}^{nir}$ are image vectors that are $N \times 1$ in size, the matrix $\mathbf{R}_i$ has dimensions $m^2 \times N$. In (4), $\mathbf{W}$ is a diagonal matrix consisting of weights that are inversely proportional to the distance between the center pixel location $i$ and its neighboring pixel location. Vector $\boldsymbol{\alpha}_i^T = [\alpha_{i,1} \ \alpha_{i,2}]$ contains two elements indicating the slope and bias, respectively. Therefore, the data fidelity term $\left\| \mathbf{W}^{1/2}(\mathbf{p}_i - \mathbf{Q}_i \boldsymbol{\alpha}_i) \right\|^2$ from (4) can be regarded as a linear mapping. In other words, the near-infrared luminance patch $\mathbf{R}_i \mathbf{x}^{nir}$ is mapped to the visible luminance patch $\mathbf{R}_i \mathbf{x}^{v_l}$ without any constraints. Adding a local contrast-preserving regularization term $\left\| \boldsymbol{\alpha}_i - \boldsymbol{\alpha}_i^0 \right\|^2$ prevents this. In this paper, the value of $\mu_c$ is set to 7,500.

| Test Images | Figs. 1(a) and (b) | Figs. 5(a) and (b) | Figs. 6(a) and (b) |
|---|---|---|---|
| MSE | $2.1 \times 10^{-5}$ | $4.1 \times 10^{-5}$ | $6.5 \times 10^{-5}$ |

Under regularization, $\boldsymbol{\alpha}_i^0$ is given as follows:

$$\boldsymbol{\alpha}_i^0 = \omega_1 \left[ \mathbf{x}_i^{nir} / avg(\mathbf{R}_i \mathbf{x}^{nir}) \quad 0 \right]^T + \omega_2 \left[ \mathbf{x}_i^{vi} / avg(\mathbf{R}_i \mathbf{x}^{vi}) \quad 0 \right]^T \quad (7)$$

where $avg$ denotes the averaging function. The above equation indicates that the center-pixel values from the extracted near-infrared and visible-luminance patches are divided by their respective average values and then combined linearly with weighting values that are set by the variance ratio between the two patches. The ratio of the center-pixel's brightness to the background brightness, as shown in (7), can be used as the local contrast measure [13,14]. In addition, the slope of $\alpha_{i,1}$ corresponds to local contrast in an image. Thus, the slope $\boldsymbol{\alpha}_i$ for the estimated linear mapping $\mathbf{Q}_i \boldsymbol{\alpha}_i$ preserves local contrast $\boldsymbol{\alpha}_i^0$ for both the near-infrared and visible-luminance patches. The closed-form solution to (4) is given as follows:

$$\boldsymbol{\alpha}_i = \left( \mathbf{Q}_i^T \mathbf{W}_i \mathbf{Q}_i + \mu_c \mathbf{I} \right)^{-1} \left( \mathbf{Q}_i^T \mathbf{W}_i \mathbf{p}_i + \mu_c \boldsymbol{\alpha}_i^0 \right) \quad (8)$$

where $\mathbf{I}$ indicates the identity matrix. Given the estimated $\boldsymbol{\alpha}_i$, the pixel value of the newly created near-infrared luminance image at the $i$th pixel location can be obtained as follows:

$$\mathbf{x}_i^{oi} = \mathbf{x}_i^{nir} \alpha_{i,1} + \alpha_{i,2} \quad (9)$$

where $\mathbf{x}_i^{oi}$ is the newly created near-infrared luminance image with the contrast-preserving mapping. Here, $\alpha_i$ is the linear-mapping relation between the luminance images.

### C. Validity of Linear Mapping Model

In (4), the relation between the near-infrared luminance image and luminance plane of the visible color image is modeled by a linear mapping function. To check this assumption, the linear mapping model was evaluated with respect to mean square error (MSE). The image pairs of the visible color and near-infrared gray images, as shown in Figs. 1, 5, and 6, were tested. MSE is defined as $\left\| \mathbf{x}^{vi} - \mathbf{x}^{oi} \right\|^2 / N$, where $\| \cdot \|$ indicates the $l_2$-norm. The pixel range of the tested images was scaled to [0–1] and the value of $\mu_c$ in (4) was set to zero. Table I shows that the calculated MSE values are quite small, and thus we conclude that the use of the linear mapping model is valid.

According to the image acquisition model [15,16], the camera response corresponds to the integral of the product of the relative power spectral distribution of the reflected light and the spectral sensitivity function over all the wavelength regions.

Let us assume that the spectral sensitivity function used to capture the luminance plane of the visible color image can be approximated by scaling and translating the spectral sensitivity function of the near-infrared luminance image as follows:

$$S^{v}(\lambda) = \omega_1 R(\lambda) + \omega_2 G(\lambda) + \omega_3 B(\lambda) \quad (10)$$

$$\begin{aligned} \int_{\lambda} S^{v}(\lambda) L(\lambda) d\lambda &\approx \int_{\lambda} \left( \alpha S^{nir}(\lambda) + \beta \right) L(\lambda) d\lambda \\ &= \alpha \left\{ \int_{\lambda} S^{nir}(\lambda) L(\lambda) d\lambda \right\} + \beta \left\{ \int_{\lambda} L(\lambda) d\lambda \right\} \end{aligned} \quad (11)$$

where $R(\lambda)$, $G(\lambda)$, and $B(\lambda)$ denote the spectral sensitivity functions of a camera's filters that respond to the long, middle, and short visible wavelength regions. In addition, $L(\lambda)$ is the relative spectral power distribution of the reflected light, and $S^{v}(\lambda)$ and $S^{nir}(\lambda)$ correspond to the spectral sensitivity functions to capture the luminance planes of the visible color and near-infrared gray images, respectively. As shown in (10), the linear combination of $R(\lambda)$, $G(\lambda)$, and $B(\lambda)$ with weight $\omega$ generate the spectral sensitivity function $S^{v}(\lambda)$ to capture the luminance plane of the visible color image. If $S^{v}(\lambda)$ can be modeled by $\alpha S^{nir}(\lambda) + \beta$, as shown in (11), the pixel value of the near-infrared gray image, described by $\int_{\lambda} S^{nir}(\lambda) L(\lambda) d\lambda$, can be linearly mapped to the pixel value of the luminance plane of the visible color image, described by $\int_{\lambda} S^{v}(\lambda) L(\lambda) d\lambda$.

### D. Detail Layer Transfer

The proposed contrast-preserving linear mapping model enables the image appearance of the newly created near-infrared luminance plane to be similar to that of the luminance plane of the visible color image. Moreover, it can transfer the local contrast of the captured near-infrared luminance plane to the newly created near-infrared luminance plane. These results can be used to resolve the discrepancy mentioned in the introduction. Furthermore, the pixel saturation of the visible color image, as shown in Fig. 1(b), can be solved. However, when mapping details of the near-infrared luminance plane to the luminance plane of the visible color image, some details can be lost. To address this issue, the detail layer of the newly created luminance plane is modified as follows:

$$\Delta \mathbf{x}^{oi} = \mathbf{x}^{oi} - \mathbf{x}^{b,oi} \quad \text{and} \quad \Delta \mathbf{x}^{nir} = \mathbf{x}^{nir} - \mathbf{x}^{b,nir} \quad (12)$$

$$\min_{\Delta \mathbf{x}} \left\{ \mu_d \left\| \Delta \mathbf{x} - \Delta \mathbf{x}^{oi} \right\|^2 + \sum_{j=1}^{2} \left| (\Delta \mathbf{x} \otimes f^j) - (\Delta \mathbf{x}^{nir} \otimes f^j) \right| \right\} \quad (13)$$

$$\mathbf{x}^{oi'} = \mathbf{x}^{b,oi} + \Delta \mathbf{x} \quad (14)$$

where $\mathbf{x}^{b,oi}$ and $\mathbf{x}^{b,nir}$ denote the base layers of the newly created near-infrared luminance plane and the captured near-infrared luminance plane, respectively, and $\Delta \mathbf{x}^{oi}$ and $\Delta \mathbf{x}^{nir}$ indicate the corresponding detail layers. As shown in (14), the newly estimated detail layer $\Delta\mathbf{x}$ is obtained via (13) and then added to the base layer $\mathbf{x}^{b,o_l}$, thus producing the detail-enhanced version $\mathbf{x}^{o_l}$. In (13), the second term brings the gradient difference between the two detail layers closer. This detail difference constraint further enhances the details of the newly created near-infrared luminance plane. In (12), the two base layers are generated. In this paper, a nonlocal means filtering was used [17]. Other filtering, e.g., bilateral filtering could be considered [18]. Equation (13) can be solved by the alternating minimization technique [6-8], where the value of $\mu_d$ is set to 200.

### E. Color Transfer

In this section, we introduce how to create colors in the newly created infrared luminance image $\mathbf{x}^{o_l}$. Because the relation between the visible color and near-infrared luminance planes has been established, the unknown colors for the newly created near-infrared luminance image $\mathbf{x}^{o_l}$ can be derived as follows:

$$\mathbf{x}_i^{o'_{c,1}} = \mathbf{x}_i^{v_{c,1}} / \alpha_{i,1} \quad \text{and} \quad \mathbf{x}_i^{o'_{c,2}} = \mathbf{x}_i^{v_{c,2}} / \alpha_{i,1} \quad (15)$$

where $\mathbf{x}_i^{v_{c,1}}$ and $\mathbf{x}_i^{v_{c,2}}$ indicate the two chrominance planes of the visible color image defined in the decorrelated color space [11]. Thus, the above equation indicates that the unknown chrominance planes $\mathbf{x}_i^{o'_{c,1}}$ and $\mathbf{x}_i^{o'_{c,2}}$ for the newly created infrared luminance image $\mathbf{x}^{o_l}$ can be obtained by dividing the chrominance planes for the visible color image by $\alpha_{i,1}$, the mapping relation. Equation (15) is derived from the contrast-preserving linear mapping—i.e., by $\mathbf{x}_i^{nir}\alpha_{i,1} \approx \mathbf{x}_i^{v}$, which reveals that the unknown chrominance planes for the newly created near-infrared gray image can be defined as the contrast-enhanced version of the chrominance planes for the visible color image. Therefore, by combining the luminance plane $\mathbf{x}^{o_l}$ with the chrominance planes $\mathbf{x}_i^{o'_{c,1}}$ and $\mathbf{x}_i^{o'_{c,2}}$, the proposed method not only preserves the local contrasts and details of the near-infrared gray image, but also transfers the colors from the visible color image to the newly created infrared luminance image.

### F. Proposed Color Transfer vs. Chroma Mapping

The proposed color transfer, as given in (15), can be regarded as a chroma mapping [19]. In the CIELAB color space, the chroma value of the newly created colored image at the $i$th pixel location can be defined as:

$$C_i^* = \sqrt{\left(\mathbf{x}_i^{v_{c,1}} / \alpha_{i,1}\right)^2 + \left(\mathbf{x}_i^{v_{c,2}} / \alpha_{i,1}\right)^2}$$

$$= \frac{1}{\alpha_{i,1}} \sqrt{\left(\mathbf{x}_i^{v_{c,1}}\right)^2 + \left(\mathbf{x}_i^{v_{c,2}}\right)^2} = s_i \sqrt{\left(\mathbf{x}_i^{v_{c,1}}\right)^2 + \left(\mathbf{x}_i^{v_{c,2}}\right)^2} \quad (16)$$

where $C_i^*$ indicates the chroma value. The above equation reveals that the proposed color transfer model linearly changes the chroma value of the captured visible color image $\mathbf{x}_i^{v}$ according to the inverse number of the local contrast $\alpha_{i,1}$. If $1/\alpha_{i,1}$ is generalized to a scaling factor $s_i$, (16) becomes the same as space-variant chroma mapping. Therefore, it is expected that the effectiveness of the proposed color transfer will be similar to that of the chroma mapping, leading to a change in the chroma.

## IV. PROPOSED NEAR-INFRARED DENOISING

In dim lighting conditions, visible color images can contain noise. Even in this case, the proposed approach for near-infrared coloring can be directly applied to the captured image pair of the noisy visible color image and near-infrared gray image; only an initial denoising is required. In other words, the proposed coloring method can be directly applied to the image pair of the denoised noisy visible color image and near-infrared gray image. The initial denoising can remove fine details in the visible color image along with the noise. However, the proposed contrast-preserving mapping can transfer the details of the captured near-infrared gray image to the newly created near-infrared gray image. Therefore, the newly created near-infrared gray image can be both noise-free and detail-preserved. Moreover, the discrepancy between the noisy visible color image and near-infrared gray image, which is no less a critical issue than the noise issue, can be resolved via the proposed contrast-preserving mapping.

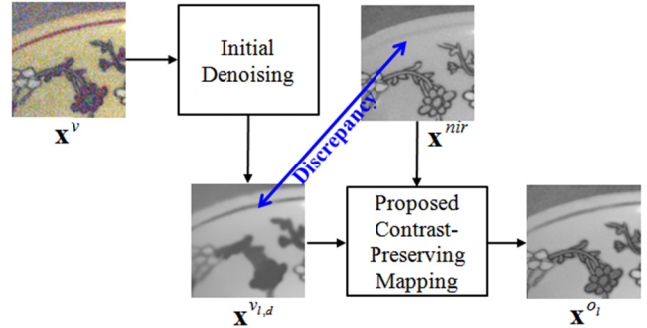

Fig. 3. Proposed scheme for near-infrared denoising.

Fig. 3 shows how the proposed near-infrared coloring method can deal with the noise and discrepancy issues. First, to remove the noise in the visible color image, an initial denoising is performed. At this time, it is recommended that the noise should be completely removed, as shown in the denoised gray image $\mathbf{x}^{v_l,d}$. Second, to restore the image structure of the denoised gray image $\mathbf{x}^{v_l,d}$, the proposed contrast-preserving mapping is used. The proposed contrast-preserving mapping can transfer details from the captured near-infrared gray image $\mathbf{x}^{nir}$ to the denoised gray image $\mathbf{x}^{v_l,d}$, as shown in the newly created near-infrared gray image $\mathbf{x}^{o_l}$. Moreover, the discrepancy between $\mathbf{x}^{v_l,d}$ and $\mathbf{x}^{nir}$ that can be found in the red lines near the brim can be resolved at the same time. The unknown colors can then be added to the newly created near-infrared gray image $\mathbf{x}^{o_l}$ according to the color transfer model in (15). Therefore, the proposed near-infrared denoising method is closer to the coloring method. This is because details and colors are transferred to the denoised gray image $\mathbf{x}^{v,d}$. This is the main difference between the proposed near-infrared denoising and conventional-image-pair-based fusions via gradient regularization [3,6,7], weighted least squares [2], and multiresolution [1,9,10] methods.

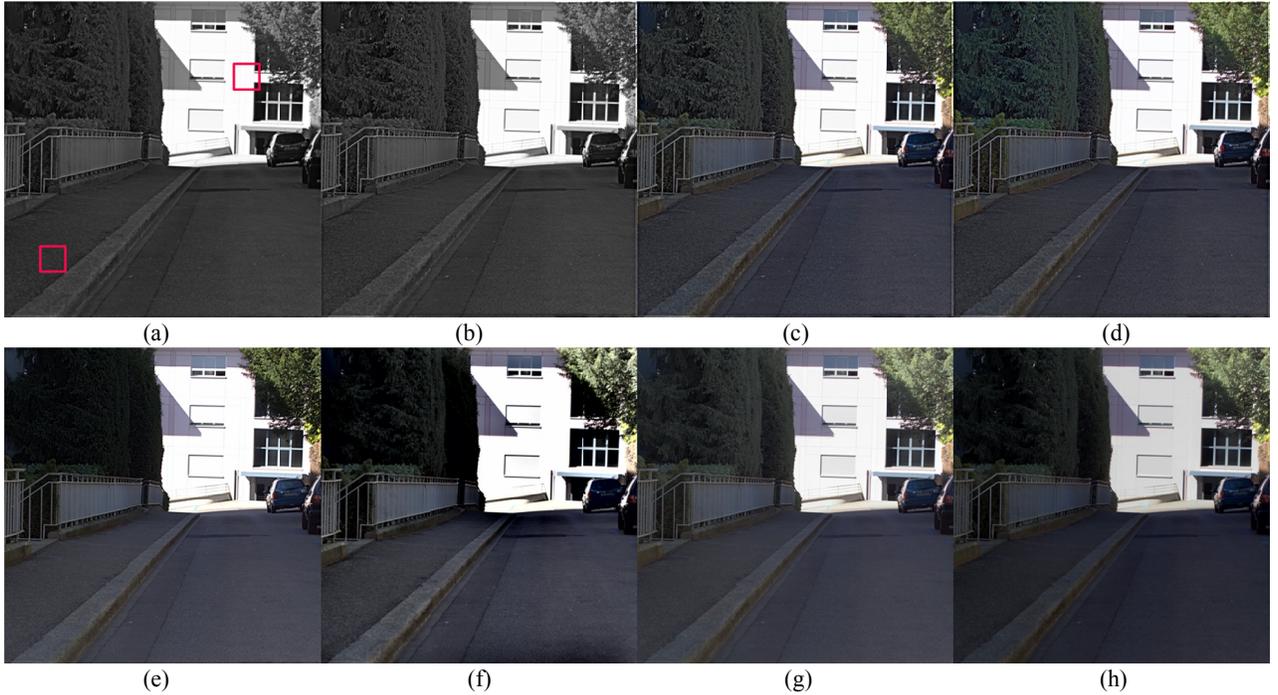

(a)  (b)  (c)  (d)

(e)  (f)  (g)  (h)

Fig. 4. Experimental results: (a) newly created infrared gray image using the proposed contrast-preserving mapping, (b) the same image with details enhanced using the proposed detail layer transfer method, (c) detail enhanced image colored by the naive method, (d) detail enhanced image colored by the proposed color-transfer method, (e) new infrared gray image colored by the gradient regularization approach ( $\mu_g = 10^3$ ), (f) new infrared gray image colored by the gradient regularization approach ( $\mu_g = 1$ ), (g) new infrared gray image colored by the multiresolution approach, (h) new infrared gray image colored by the dehazing technique [4].

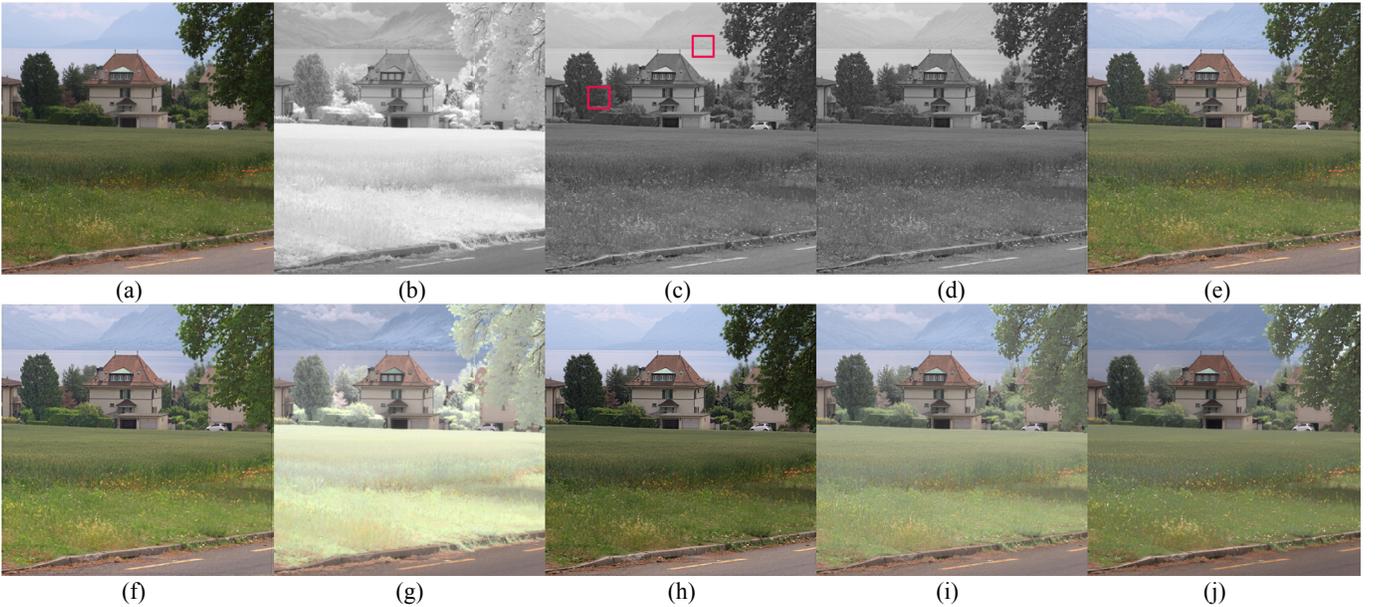

(a)  (b)  (c)  (d)  (e)

(f)  (g)  (h)  (i)  (j)

Fig. 5. Experimental results: (a) visible color image, (b) near-infrared gray image, (c) newly created infrared gray image using the proposed contrast-preserving mapping, (d) newly created infrared gray image with details enhanced using the proposed detail layer transfer method, (e) detail enhanced image colored by naive method, (f) detail enhanced image colored by the proposed color-transferring method, (g) new infrared gray image colored by the naive method, (h) new infrared gray image colored by the gradient regularization approach ( $\mu_g = 10^3$ ), (i) new infrared gray image colored by the multiresolution approach, and (j) new infrared gray image colored by the dehazing technique [4].

## V. EXPERIMENTAL RESULTS

### A. Visual Quality Comparison

To verify the visual effectiveness of the proposed coloring method, test images that include high contrast and detail were chosen from the near-infrared image database [1], as shown in Figs. 4, 5, 6, and 7. Also, to compare the performance of the proposed method, the conventional image fusion techniques such as the gradient regularization [3,6,7], multiresolution [1, 9,10], and multiscale tone and detail manipulation [4] were adopted because the near-infrared coloring can be considered as an image fusion problem. To reiterate the points mentioned above, the proposed coloring method is good at representing local detail, contrast, and realistic colors. As shown in Fig. 4(a), the newly created near-infrared luminance image, rendered using (4), preserved detail in the tree regions. Furthermore, the tree regions that have been darkened in the visible color image because of the low dynamic range of the camera, as depicted in coloring method are better than those resulting from conventional methods, as shown in Figs. 4(e)-(h). The difference between Figs. 4(e) and 4(f) depends on the value of $\mu_g$, which is defined in (1). A lower value enables the preservation of the edges from the near-infrared gray image. Despite an increase in the edges of the tree regions, the colors become distorted, as shown in Fig. 4(f). The multiresolution approach, shown in Fig. 4(g), preserves the details but results in unnatural colors. The dehazing technique [4] cannot solve the discrepancy problem. The brightness of the tree's regions is still dark and some lines on the wall are almost removed, as shown in Fig. 4(h). The statistical approach [11] was not compared in this paper. This is because its visual qualities were not better than those of the gradient regularization, dehazing, and multiresolution approaches. In particular, the colors it produced were unnatural and oversaturated.

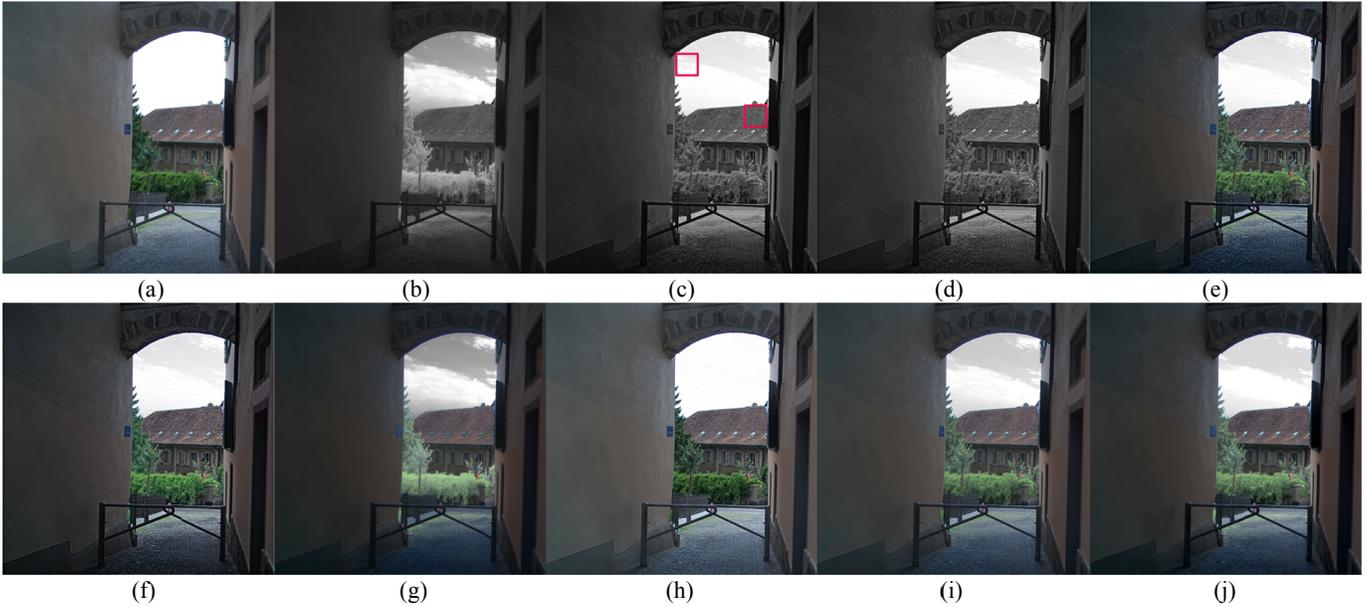

Fig. 6. Experimental results: (a) visible color image, (b) near-infrared gray image, (c) newly created infrared gray image with the proposed contrast-preserving mapping, (d) newly created infrared gray image with details enhanced using the proposed detail layer transfer method, (e) detail enhanced image colored by the naive method, (f) detail enhanced image with the proposed color-transferring method, (g) new infrared gray image colored by the naive method, (h) new infrared gray image colored by the gradient regularization approach ( $\mu_g = 10^3$ ), (i) new infrared gray image colored by the multiresolution approach, and new infrared gray image colored by the dehazing technique [4].

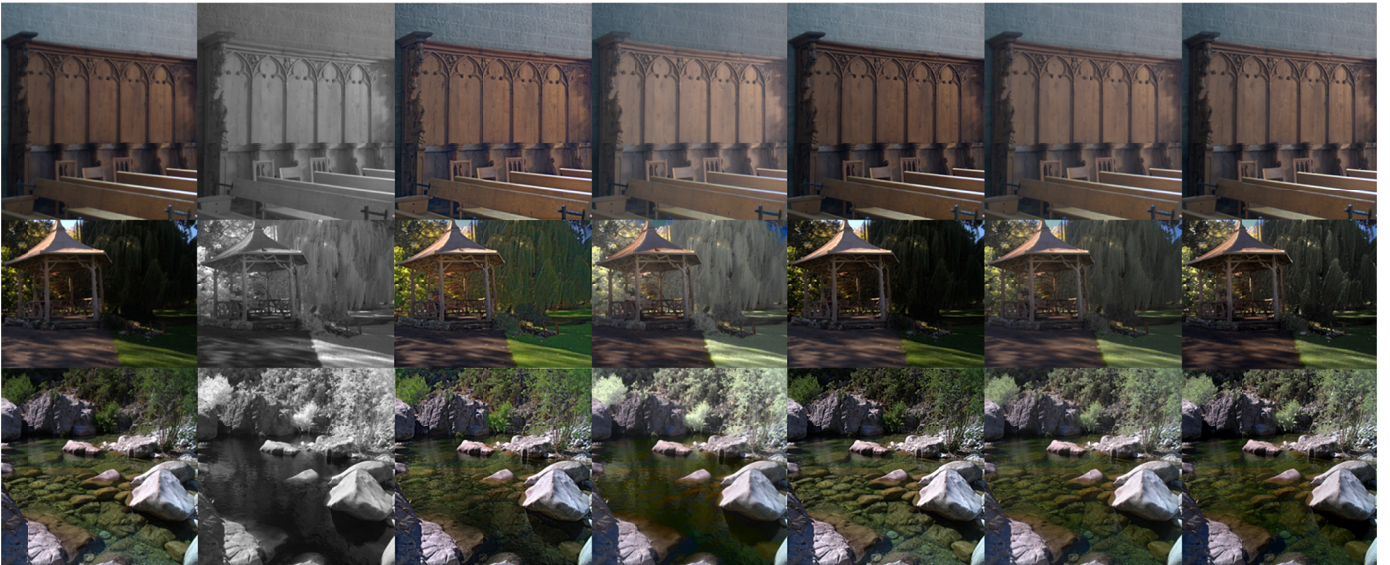

Fig. 7. Experimental results: visible color images (first column), near-infrared gray image (second column), and images colored by the proposed method (third column), naive method (fourth column), gradient regularization approach (fifth column), multiresolution approach (sixth column), and dehazing technique [4] (seventh column).

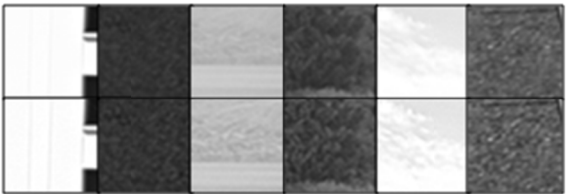

Fig. 8. Detail comparison: before detail layer transfer (upper row) and after detail layer transfer (bottom row).

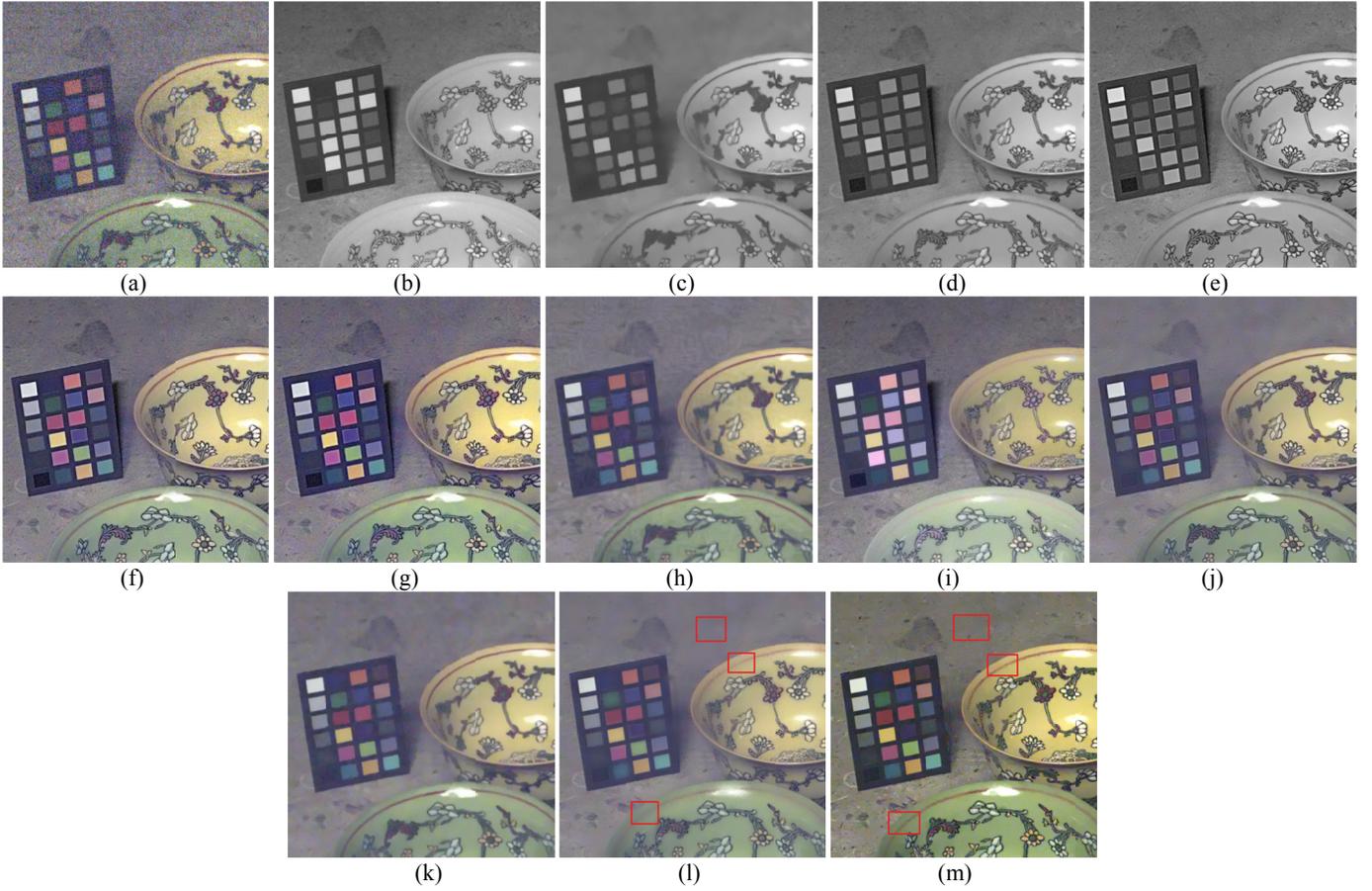

Fig. 9. Experimental results: (a) noisy visible color image, (b) near-infrared gray image, (c) image denoised via nonlocal means filtering, (d) newly created near-infrared gray image via the proposed contrast-preserving mapping, (e) newly created near-infrared gray image with details enhanced using the proposed detail layer transfer method, (f) image colored using the proposed color transfer method, (g) chroma-enhanced version of the Fig. 9(f), (h) image denoised using BM3D [20], (i) image colored by the naive method, (j) image denoised using the gradient regularization approach, (k) image denoised using the weighted least squares approach, (l) image denoised using guided image filtering [21], and (m) image denoised using scale map method [22].

Fig. 1(b), can be recovered with higher contrast. However, detail can be lost when mapped from the near-infrared gray image into the luminance plane of the visible color image. For example, the lines on the wall in the highlight regions were almost lost. However, this drawback can be corrected via the detail layer transfer of Eqs. (12)–(14). In Fig. 4(b), the lines on the wall are restored. In addition, the details of the tree's leaves and road surfaces were also enhanced. For comparison, magnified versions of the red rectangles marked in Figs. 4(a), 5(c), and 6(c) are shown in Fig. 8. From these results, it can be concluded that the detail layer transfer improves the amount of information and detail description. The naive coloring method of combining the chrominance planes of the visible color image with the newly created near-infrared luminance image resulted in improved colors, as shown in Fig. 4(c). The colors are more natural than those of Fig. 1(c), indicating that the discrepancy problem can be resolved by the proposed contrast-preserving mapping. However, the colors appear desaturated. The proposed color-transfer method, as defined in (15), improved the colors by estimating the contrast-enhanced version of the chrominance planes for the visible color image, as shown in Fig. 4(d). In addition, the colors produced with the proposed

Similar effects can found in Figs. 5 and 6. It can be seen that the discrepancy problem occurs, especially in the grass region of Figs. 5(a) and (b) and the cloud region of Figs. 6(a) and (b). Thus, unnatural colors were produced by the naive coloring method, as shown in Figs. 5(g) and 6(g). In Figs. 5(f) and 6(f), the proposed method produced better colors than the naive, gradient regularization, multiresolution, and dehazing approaches. However, the conventional methods have their own merits. The gradient regularization approach is good at representing edges, as shown in the grass region of Fig. 5(h) and the bush region of Fig. 6(h). In contrast, the multiresolution approach is good at increasing the amount of visual information, especially in the distant mountain region of Fig. 5(i) and the cloud region of Fig. 6(i). The disadvantage to using the gradient regularization approach is that it often omits visual information, for example, the clouds in the sky have been removed, as shown in Fig. 6(h). The disadvantage of the multiresolution approach is that the colors it produces are often unrealistic, as shown in Fig. 5(i), and the strength of edges are relatively weak when compared to the gradient regularization and proposed method, as in the roof regions of Fig. 6(i). The advantage and disadvantage of the dehazing approach [4] are similar to those of the multiresolution approach, as checked in Figs. 5(j) and 6(j). This is because the dehazing method is based on the multiscale tone and detail manipulation. Thus, both the proposed and conventional methods have their own merits and demerits. However, the proposed method is more effective at expressing the color contrasts than conventional methods. In addition, fine details and natural colors can be obtained simultaneously. This is possible because the proposed coloring method adopts three kinds of transfer: contrast, detail, and color, as shown in the block-diagram of Fig. 2. In addition, the proposed method lays more emphasis on near-infrared coloring, whereas the conventional methods focus on fusing the images for the improvement of edge representation and information preservation. This is the main difference between the proposed and conventional methods. For a more visual comparison, additional image results are provided in Fig. 7. Moreover, a few visible/near-infrared image sets including thin lines, high texture content, or a combination of high texture and flat areas are tested and then provided in Supplement. The performance results are similar to Figs. 4-7. All the resulting images can be downloaded from our website for the paper (https://sites.google.com/site/ changhwan76son/).

TABLE II
DEFINITIONS OF THE FOUR MEASURES

| |
|---|
| $CF = \sigma_{ab} + 0.94u_C$, where $\sigma_{ab}$ and $u_C$ are related to the standard deviation of the chrominance planes and mean value of the chroma image, respectively. |
| $SF = \sqrt{R_F^2 + C_F^2}$, where $R_F$ and $C_F$ indicate the averaged numbers of the vertical and horizontal edges, respectively. |
| $EN = -\sum_{i=1}^{L} \mathbf{h}(i) \log \mathbf{h}(i)$, where $\mathbf{h}$ indicates the normalized histogram of a colored image. |
| $CT = \frac{1}{N} \sum_{i=1}^{N} \left| \mathbf{x}_i^{p'} - \mathbf{x}_i^{p} \otimes \mathbf{k} \right|$, where $\mathbf{k}$ indicates a Gaussian filter. |

TABLE III
QUANTITATIVE EVALUATION

| Test Images | Methods | CT | EN | SF | CF |
|---|---|---|---|---|---|
| | Naïve method | 12.949 | 14.623 | 15.749 | 11.891 |
| | Gradient regularization | 17.375 | 14.770 | 23.367 | 11.893 |
| Fig. 4 | Multi-resolution | 14.516 | 14.761 | 18.68 | 11.936 |
| | Dehazing method [4] | 13.070 | 14.552 | 16.863 | 12.394 |
| | Proposed method | **19.046** | **14.846** | **23.675** | **13.732** |
| | Naïve method | 13.828 | 15.625 | 15.219 | 28.583 |
| | Gradient regularization | 19.503 | **15.742** | 22.5 | 31.487 |
| Fig. 5 | Multi-resolution | 16.356 | 15.497 | 18.711 | 29.781 |
| | Dehazing method [4] | 16.112 | 15.261 | 19.367 | 30.946 |
| | Proposed method | **20.111** | 15.704 | **23.445** | **38.373** |
| | Naïve method | 8.512 | 14.322 | 9.797 | **11.317** |
| | Gradient regularization | 12.991 | 14.302 | 17.159 | 11.052 |
| Fig. 6 | Multi-resolution | 10.603 | 14.423 | 13.54 | 11.133 |
| | Dehazing method [4] | 10.857 | 14.314 | 12.999 | 11.208 |
| | Proposed method | **13.741** | **14.555** | **17.845** | 10.366 |
| | Naïve method | 10.657 | 14.826 | 10.98 | 16.397 |
| | Gradient regularization | 12.909 | 15.113 | 14.128 | 17.604 |
| Fig. 7 (First row) | Multi-resolution | 10.995 | 14.945 | 11.616 | 16.914 |
| | Dehazing method [4] | 12.157 | 15.033 | 12.447 | 17.114 |
| | Proposed method | **14.656** | **15.138** | **15.787** | **21.759** |
| | Naïve method | 19.025 | **15.410** | 17.031 | 17.164 |
| | Gradient regularization | 19.464 | 14.895 | 17.122 | 17.483 |
| Fig. 7 (Second row) | Multi-resolution | 18.158 | 15.083 | 16.798 | 17.685 |
| | Dehazing method [4] | 12.874 | 15.105 | 15.962 | 17.925 |
| | Proposed method | **23.630** | 15.205 | **20.238** | **30.029** |
| | Naïve method | 24.637 | **15.944** | 24.751 | 21.608 |
| | Gradient regularization | **33.450** | 15.922 | **36.554** | 22.587 |
| Fig. 7 (Third row) | Multi-resolution | 27.209 | 15.893 | 28.758 | 22.623 |
| | Dehazing method [4] | 25.785 | 15.795 | 27.206 | 22.702 |
| | Proposed method | 31.207 | 15.889 | 34.068 | **24.579** |
| | Naïve method | 14.935 | 15.125 | 15.588 | 17.827 |
| | Gradient regularization | 19.282 | 15.124 | 21.805 | 18.684 |
| AVG. | Multi-resolution | 16.306 | 15.100 | 18.017 | 18.345 |
| | Dehazing method [4] | 15.143 | 15.010 | 17.474 | 18.715 |
| | Proposed method | **20.399** | **15.223** | **22.510** | **23.190** |

Fig. 9 shows the experimental results for the image pair captured in dim lighting [2]. In Figs. 9(a) and (b), it is clear that the visible color image contains noise and discrepancy occurred, especially for the chart's patches and red lines near the brim of the bowl. Fig. 9(c) shows the initially denoised visible color image via nonlocal means filtering [17]. In this figure, the details have clearly been removed. However, this detail loss can be restored by applying the proposed contrast-preserving mapping and detail layer transfer, as shown in Figs. 9(d) and (e). In Fig. 9(d), it is clear that the newly created near-infrared gray image preserves its detail thanks to the proposed contrast-preserving mapping. Moreover, the discrepancy problem can be resolved. For example, the red line on the bowl that was removed in the near-infrared gray image of Fig. 9(b) is restored. In Fig. 9(e), the use of the detail layer transfer leads to an improvement in the detail description. Fig. 9(f) shows the colored image with the proposed color transfer method and Fig. 9(g) is the chroma-enhanced version of the Fig. 9(f). That is, the chroma mapping, as shown in (16), is applied to the Fig. 9(f). The scale factor, $s_i$, is set with 1.2. In the case of the noisy visible and near-infrared image pairs, the values of the estimated linear mapping relation, $\alpha_i$, in (4), can be shrunk. This is because the used denoised luminance image has already lost its contrast and edge, as shown in Fig. 9(c), which can lead to the decrease in chroma. Different from other resulting images, as shown in Figs. 9(h)-(m), the colored image with the proposed denoising method has little noise and color distortions. Thus, the chroma mapping can improve the colorfulness without noise amplification, as shown in Fig. 9(g). BM3D [20] is known as one of the state-of-the-art denoising methods. However, its visual quality is poor, as shown in Fig. 9(h). It is guessed that the main reason is due to the non-Gaussian noise in the captured visible color image. The image-pair-based denoising methods based on the weighted least squares [2] and gradient regularization [3,6,7] can produce better-resulting images than those obtained from the BM3D. This is possible because the clean near-infrared image is used as a guidance image [2]. Figs. 9(l) and (m) show the denoised images with the guided image filtering [21] and scale map method [22], respectively. As shown in the red boxes of Fig. 9(l), the lines cannot be restored. Also, the background textures are almost removed. This indicates that the guided image filtering is not suitable for resolving the discrepancy problems. Similarly, the scale map method cannot clearly restore the lines, as shown in red boxes. The edges of the brim of the bowl are broken. Also, the background colors are different from those of the captured visible color image. Based on the comparison of the resulting images, the visual quality of the proposed method is better than those of the conventional methods. This is possible because the proposed denoising method is based on both detail and color transfers, as discussed in Section IV. In other words, the proposed denoising method adds details and colors to the denoised luminance image of Fig. 9(c). This is the main difference between the proposed denoising method and the conventional image fusion methods [2,3,21,22]. From these results, it can be concluded that the proposed near-infrared coloring method is effective at removing noise of the captured visible color images in dim lighting condition.

Also, note that in Fig. 9, the weighted least squares and gradient regularization approaches utilized both the noisy visible color image and near-infrared gray image. In other words, the denoised version of the noisy visible color image via the nonlocal means denoising is used as a guidance image in the weighted least squares approach [2]. Also, the gradient regularization approach used the denoised image to model the data-fidelity term, as shown in (1). Therefore, the weighted least squares and gradient regularization approaches can be considered as multi-frame denoising methods as well.

### B. Quantitative Evaluation

To evaluate the performance of the coloring methods, four types of measures: colorfulness (CF) [23], spatial frequency (SF) [24], entropy (EN) [25], and contrast (CT) [13,14] were chosen. These measures are defined in Table II. Further details about their variables can be found in the related references. These four measures have been used for the quantitative evaluation of other types of image pairs [23,25]. In this paper, the CF measure is used to perceptually quantify the colorfulness in a colored image and the SF measure is used to investigate edge preservation. The EN and CT measures quantify the amount of information and local contrast in a colored image, respectively. At least one of the contrast, edge, color, or visual information can be lost after applying image fusion including the near-infrared coloring. Therefore, it is necessary to measure how well the contrast, edge, color, and visual information are represented on the fused or colored images. For this reason, the four measures were adopted to evaluate the near-infrared coloring. Image quality measures that require a reference image [25] were not considered in this paper. This is because both visible color and near-infrared gray images can contain image quality degradation, and thus neither of those images can be used as the reference image. For example, as shown in Figs. 1(a) and (b), the visible color image includes contrast loss and the near-infrared gray image has no color. For this reason, reference image fusion metrics, such as the root mean square error (RMSE) or peak signal to noise ratio (PSNR), were not considered. Even though there are some blind image quality evaluations where natural statistics of edges, colors, or contrasts are modeled to predict distortion levels [26], image enhancement algorithms such as sharpening, coloring, dehazing, or etc., lead to inevitable modification in the natural statistics. Thus, the blind image quality evaluation is not appropriate for near-infrared coloring.

As expected, the naive coloring method yielded the lowest average scores, as shown in the lowest partition of Table III. However, the EN score is relatively high, especially for the test images located in the second and third rows of Fig. 7. The EN measure has the highest score when a discrete random variable has a uniform probability distribution [27], i.e., when the histogram of an image is uniform. Because these two test images have a wide range of pixel intensities, they obtained the highest EN scores. The gradient regularization approach is good at representing edges, and thus it obtains higher SF scores, especially for the test image of Fig. 7 (third row). A small value of $\mu_g$ in (1) can increase the SF scores, however color distortion can occur, as shown in Figs. 4(e) and (f). In contrast, the proposed method obtained the highest average scores for all evaluations, as shown in the lowest partition of Table III. This is because of the three types of transfer: contrast-preserving mapping, detail layer transfer, and color transfer. The contrast-preserving mapping preserves the local contrast of the near-infrared gray image and the color transfer increases the chrominance range of the visible color image. The detail layer transfer increases the strength of details and edges. The perceptual meaning of the CF score (the final column of Table III) was introduced in [23]. For example, a score that lies between 8 and 18 indicates that a tested image is slightly colorful and a score between 32 and 43 indicates that a tested image is quite colorful. Note that the proposed method obtains higher CF scores for the colorful images of Figs. 5 and 7 (second row). In Table III, the quantitative evaluations of the multiresolution and dehazing approaches are not satisfactory. However, specific areas in the whole images, e.g., sky regions, can be well rendered. This indicates that those two approaches are more appropriate for dehazing [4], not the near-infrared coloring [1].

## VI. Conclusion

In this paper, a new method is presented to transfer colors from a visible color image to a near-infrared gray image without losing details and contrast. Specifically, based on a new contrast-preserving mapping model, the proposed contrast-preserving mapping method can solve the discrepancy problem in brightness and image structures between the visible and near-infrared images. Visual information loss during the contrast-preserving mapping can be compensated with the transfer of image details, thereby restoring thin lines and improving fine details. In addition, dehazing effects can be simultaneously obtained through the proposed contrast-preserving mapping model and the detail transfer procedure. Moreover, the proposed color-transfer technique generates realistic colors that can be added to the newly created near-infrared gray image by applying the space-variant chroma mapping to the chrominance planes of the captured visible color image. The proposed approach for the near-infrared coloring can also be used to remove noise in visible color image captured in dim lighting condition. Experimental results show the effectiveness and superiority of the new method in comparison with the commonly used methods.